\documentclass[runningheads]{llncs}
\usepackage{graphicx}
\usepackage{todonotes}
\usepackage{subfigure}
\usepackage{amsfonts}
\usepackage{multirow}
\usepackage{amsmath}
\usepackage{array}
\usepackage{float}
\usepackage{xurl}

\begin{document}

\title{Brazilian Portuguese Speech Recognition Using Wav2vec 2.0}
% \thanks{Supported by organization x.}

\author{
Lucas Rafael Stefanel Gris \inst{1}\thanks{~Corresponding author: gris at alunos utfpr dot edu dot br}\orcidID{0000-0002-2099-5004}, \\
Edresson Casanova\inst{2}\orcidID{0000-0003-0160-7173}, \\
Frederico Santos de Oliveira\inst{3}\orcidID{0000-0002-5885-6747}, \\
Anderson da Silva Soares\inst{4}\orcidID{0000-0002-2967-6077}, \\
Arnaldo Candido Junior\inst{1}\orcidID{0000-0002-5647-0891}
}

% \author{
% omitted
% }

\authorrunning{Gris et al.}
% \authorrunning{omitted.}

% \institute{omitted}
\institute{Federal University of Technology - Paraná, Medianeira, Brazil
\and
 University of São Paulo, São Carlos, Brazil
\and
Federal University of Mato Grosso, Mato Grosso, Brazil
\and
  Federal University of Goias, Goiânia, Brazil
%   \email{gris at alunos dot utfpr dot edu dot br %\linebreak
%   }
}


\maketitle              % typeset the header of the contribution

\begin{abstract}
Deep learning techniques have been shown to be efficient in various tasks, especially in the development of speech recognition systems, that is, systems that aim to transcribe an audio sentence in a sequence of written words. Despite the progress in the  area, speech recognition can still be considered difficult, especially for languages lacking available data, such as Brazilian Portuguese (BP). In this sense, this work presents the development of an public Automatic Speech Recognition (ASR) system using only open available audio data, from the fine-tuning of the Wav2vec 2.0 XLSR-53 model pre-trained in many languages, over BP data. The final model presents an average word error rate of 12.4\% over 7 different datasets (10.5\% when applying a language model). According to our knowledge, the obtained error is the lowest among open end-to-end (E2E) ASR models for BP.

\end{abstract}

%\arnaldo{1. dizer que somos modelo aberto}

%\arnaldo{2. explicar o que faltou das letrinhas nas equações 2 e 3}

%\arnaldo{3. passar link para Anderson, avisar Fred e Edresson que fechamos}

%\arnaldo{4. não esquecer de comentar a autoria por causa da blind revision}

% nosso resultado é melhor que quintanilha, mas os datasets não são diretamente comparáveis. Acreditamos SOTA.

%\lucas{PROPOR dia 25/10}
%\fred{Comentários Fred}
%\anderson{Comentários Anderson}

\section{Introduction}

Speech is one of the most natural ways of human communication, and the development of systems, known as Automatic Speech Recognition (ASR) Systems, capable of transcribing speech automatically have shown great importance and applicability in various scenarios, as in personal assistants, tools for customer attendance and other products \cite{karpagavalli2016review,goodfellow,yu2016automatic}.
The task to transcribe speech can be understood as a mapping of an acoustic signal containing speech to a corresponding sequence of symbols intended by the speaker \cite{goodfellow}. The research in the field started with the recognition of spoken digits \cite{davis1952automatic} and has shown great advances recently with the use of E2E  deep learning models, especially  for the English language.

% ASR systems can be built using different learning approaches. The most common is probably the supervised version, which is based on training over a dataset containing speech data with its respective labels.
%to train a model with the objective to predict transcriptions of a given audio. Unsupervised techniques can be used as mitigating the lack of available data. 
%Usually supervised learning requires aligned labels with the respect to each speech audio at different levels, such as phones, words or sentences. Unsupervised learning can use unaligned text and speech as part of the training process.
% Besides supervised, other techniques can be used as well. In particular, self-supervised learning has gained interest. This technique usually consists of using the unlabeled speech audio for training a model that is able to learn representations from the samples themselves. 
%These representations can be learned by predicting or reconstructing missing and masked parts of each sample or by solving a contrastive task, where the model needs to differentiate distinct samples from false examples \cite{baevski2019vq}. 

Despite its progress in the area, the development of robust ASR models for languages other than English can still be considered a difficult task, mainly because state-of-the-art (SOTA) models usually needs many hours of annotated speech for training to achieve good results \cite{amodei2016deep,quintanilha2020open}. This can be a challenge for some languages, such as BP, that has just a fraction of open resources available, if compared to the English language \cite{neto2011free,neto2008spoltech}. 
In general, the accuracy of Portuguese ASR models are far from  researches for more popular languages \cite{AGUIARDELIMA2020101055}.
%To overcome these challenges, one could manually transcribe the amount of data needed to produce robust models, which is an expensive and time-consuming process. Another option is the use of modern techniques such as self-supervised learning, which has shown great results in the development of models where there is a lack of available labeled data. 
To overcome these challenges, self supervised  learning can be used to learn representations for posterior use in a fine-tuning process using less labeled data \cite{baevski2020Wav2vec}.

Wav2vec 2.0 \cite{baevski2020Wav2vec} is a modern deep learning architecture proposed for ASR that employs self supervision for speech representation learning.
% The authors shows that Wav2vec 2.0 allows the training of a robust ASR model in English using only 10 minutes of labeled data, which outperformed several SOTA models. This demonstrates a good alternative for languages with few resources available.
%Although it achieves good results in English with few labeled data, it is not clear if the same results could be obtained when using the same pre-training in other languages. 
%In other words, it is possible that monolingual training could led to poor results when using this model for a language different than the original. Fortunately, 
Although the original work focused on the English language,  \cite{conneau2020unsupervised} pre-trained a new version, called Wav2vec 2.0 XLSR-53, using 56k hours of speech audio of 53 different languages, including BP.
%Besides masking and contrastive learning, this model jointly learned quantized representations across multiple languages, producing a cross-lingual model that significantly outperforms monolingual pre-training and enables a single multilingual speech recognition model which is competitive to the individual ones.
This work uses the model pre-trained by \cite{conneau2020unsupervised} 
%in many languages 
to build an ASR for BP, using only open available data. The resulting fine-tuned model for BP and code of this work is publicly available\footnote{\url{https://github.com/lucasgris/wav2vec4bp}}. 

This work is organized as follows: Section \ref{sec:wav2vec} explains the Wav2vec 2.0 model, Section \ref{sec:asr-br} provides details about the best available open model for ASR in BP, Section \ref{sec:methods} discuss the proposed method and Section \ref{sec:results}
shows and dicusses the obtained results.
% demonstrates the obtained results. Section \ref{sec:comparison-br} discusses some comparisons with the best open BP ASR model and similar works. 
Finally, Section \ref{sec:conc} presents conclusions of the work.

\section{Wav2vec}\label{sec:wav2vec}

Wav2vec 2.0 is an E2E  model inspired on the previous works \cite{schneider2019wav2vec,baevski2019vq}. %that uses self supervised learning based on a pre-training with speech masking to solve a contrastive task, followed by a supervised fine-tuning on transcribed speech. This model can jointly learn quantized representations across multiple languages.
The pre-training uses speech masking in order to solve a contrastive task to differentiate true quantized latent representations from a set of distractors, which allows the model to learn audio representations from speech. During fine-tuning, a projection is added to the last layer of the model containing the respective vocabulary and the model is trained in a supervised way using aligned labeled speech to perform ASR. 
%As in Vq-wav2vec, the model is pre-trained in a self-supervised way using the raw audio to learn discrete representations of speech, but it has an simpler architecture when compared to the Vq-wav2vec. 
The model is based on the idea of taking discrete speech representations directly as input to the Transformer \cite{vaswani2017attention} (called Context Network). 
%, solving the ASR as an E2E  task \cite{baevski2020Wav2vec}, 
%whereas in Vq-wav2vec a BERT\cite{devlin2018bert} model was trained using discrete representations, and then used as input to a separated acoustic model.

\begin{figure}[htp]
	\centering
	\includegraphics[width=.7\textwidth]{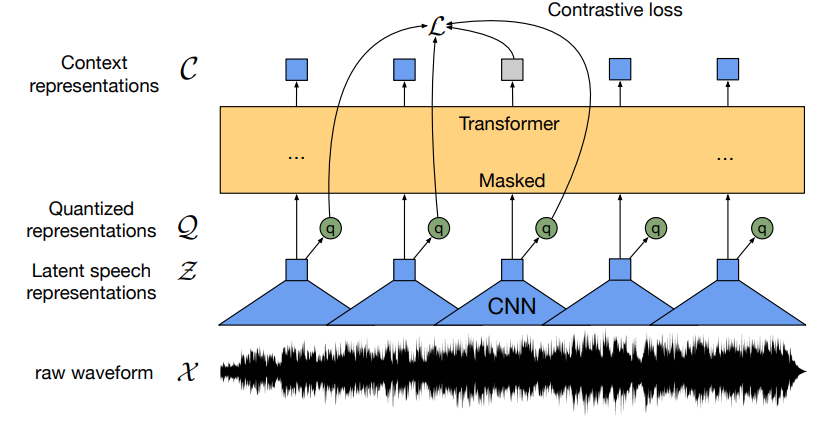}
	\caption{Wav2vec 2.0 architecture. \cite{baevski2020Wav2vec}}
\label{fig:arquitetura-wav2vec2}
\end{figure}

The architecture of the Wav2vec 2.0 is presented in Figure \ref{fig:arquitetura-wav2vec2}. 
%Similarly to original model,  Wav2vec 2.0 architecture 
It is composed by a multi-layer convolutional encoder $f: X \longmapsto Z$ which maps raw speech $X$ into latent speech representations $z_1, \ldots, z_T$ in $T$ time-steps. The convolutional blocks of the encoder consist of causal convolutions followed by layer normalization \cite{ba2016layer} and the GELU activation function \cite{hendrycks2020gaussian}. The output of the encoder is then provided to the context network $g: Z \longmapsto C$ which maps the latent representations $Z$ into contextualized representations $c_1, \ldots, c_T$. The context network follows the Transformer architecture. In this network, the positional encoding is replaced by a convolutional layer with GELU activation function and layer normalization that acts as a relative positional embedding. 

% \begin{figure}[h]
% 	\centering
% 	\includegraphics[width=.8\textwidth]{figures/wav2vec2_alt.png}
% 	\caption{Arquitetura do Wav2vec 2.0. \cite{yi2020applying}}
% \label{fig:arquitetura-wav2vec2}
% \end{figure}

During the pre-training phase, the model has the objective to learn the speech representations solving a contrastive task.
%$\mathcal{L}_{m}$. 
This corresponds to the task of identifying a true quantized representation $q_t$ from a set of false examples in a masked time-step context. 
After pre-training, a projection of $n$ targets is added to the output of the context network and the model is fine-tuned in a supervised task for speech recognition \cite{baevski2020Wav2vec}. In this phase, the model is trained using the Connectionist Temporal Classification (CTC) \cite{graves2006connectionist} as loss function. An approach similar to SpecAugment \cite{Park_2019} is also used, promoting a better generalization to the ASR model.

The authors proposed a series of experiments using two versions of the model, BASE and LARGE (the LARGE has more parameters in the context network) and varying the datasets used to pre-train and fine-tune the model. \cite{baevski2020Wav2vec} showed that it is possible to build ASR models even with few labelled data available. The experiments with 10 minutes of labeled data showed a Word Error Rate (WER) of 4.8\% and 8.2\% in both LibriSpeech test sets, namely clean and other, respectively. In this work, we use Wav2vec 2.0 XLSR-53 \cite{conneau2020unsupervised}, which is based on the LARGE architecture.

\section{ASRs for Brazilian Portuguese}\label{sec:asr-br}

The accuracy of ASR systems has been increased with the use of new technologies based on neural networks, in particular, with the development of E2E models, which achieved most of the recent state-of-the-art results  \cite{li2021recent}. Despite the development of new technologies among ASR researchers, there is still few work related to speech recognition for Brazilian Portuguese.

An important ASR baseline for BP is the work of \cite{batista18iberspeech}. The authors developed baseline ASR systems using the Kaldi toolkit \cite{povey2011kaldi} by training several acoustic and language models using various free and paid BP datasets (170h in total). Although the developed systems are not based on the use of deep learning techniques and modern E2E topologies, the results are promising, with the lowest WER being 4.75\% against LapsBM Benchmark \cite{neto2011free}.

In the context of E2E models, the works of \cite{quintanilha2017end}, \cite{quintanilha2020open} and \cite{gris2021desenvolvimento} can be highlighted as important recent advances. The work of \cite{quintanilha2017end} presented a dataset in Portuguese composed by various freely available datasets (SID, VoxForge, LapsBM) and the proprietaty dataset CSLU Spoltech \cite{schramm2006cslu}.
%The Spoltech dataset contains 8,080 sentences and 477 speakers, the Sid contains 5,777 sentences and 72 speakers, the VoxForge contains 4,090 sentences and more than 111 speakers and LapsBM contains 700 sentences and 35 speakers.
The author proposed an E2E model based on a simple architecture containing Bidirectional LSTM \cite{hochreiter1997long} layers. The obtained error was 25.13\% in the proposed test set.

More recently, \cite{quintanilha2020open} proposed a better version of the previous dataset and trained a topology based on the DeepSpeech 2 \cite{amodei2016deep}, containing two convolutional layers and five bidirectional recurrent layers. The authors merged the CETUC \cite{alencar2008lsf} dataset with the previous one, which allowed the training of deeper models. The authors also trained Language Models (LMs) based on KenLM \cite{heafield2011kenlm} for post-processing the transcriptions. The work of \cite{quintanilha2020open} presented a WER of 25.45\% in the proposed test set.

Regarding the use of the Wav2vec 2.0 architecture to build ASRs for BP, we can cite \cite{gris2021desenvolvimento} as an important contribution. In this previous work, we used a low-resource data (LapsBM dataset) to fine-tune the XLSR-53 model for the BP language. We validated and tested the model using the Common Voice dataset, which demonstrated promising results (34\% WER) even using only 1 hour to finetune the model.

\section{Datasets and Experiments}\label{sec:methods}

This section presents the datasets used to fine-tuning the Wav2vec 2.0 model for Brazilian Portuguese. The proposed datasets are:

\begin{itemize}
    \item CETUC \cite{alencar2008lsf}: contains approximately 145 hours of Brazilian Portuguese speech distributed among 50 male and 50 female speakers, each pronouncing approximately 1,000 phonetically balanced sentences selected from the CETEN-Folha\footnote{\url{https://www.linguateca.pt/cetenfolha/}} corpus;
    \item LaPS Benchmark \cite{neto2011free} \footnote{``Falabrasil -- UFPA'' (\url{https://github.com/falabrasil/gitlab-resources})} (LapsBM) is a dataset used by the Fala Brasil group to benchmark ASR systems in Brazilian Portuguese. Contains 35 speakers (10 females), each one pronouncing 20 unique sentences, totalling  700 utterances in Brazilian Portuguese;
    \item VoxForge\footnote{\url{http://www.voxforge.org/}}: is a project with the goal to build open datasets for acoustic models. The corpus contains approximately 100 speakers and 4,130 utterances of Brazilian Portuguese, with sample rates varying from 16kHz to 44.1kHz.
    \item Common Voice (CV) 7.0: is a project proposed by Mozilla Foundation with the goal to create a wide open dataset in different languages. In this project, volunteers donate and validate speech using the official site\footnote{\url{https://commonvoice.mozilla.org/pt}};
    \item Multilingual LibriSpeech (MLS) \cite{Pratap_2020}: a massive dataset available in many languages. The MLS is based on audiobook recordings in public domain like LibriVox\footnote{\url{https://librivox.org/}}. The dataset contains a total of 6k hours of transcribed data in many languages. The set in Portuguese used in this work\footnote{\url{http://www.openslr.org/94/}} (mostly Brazilian variant) has approximately 284 hours of speech, obtained from 55 audiobooks read by 62 speakers;
    \item Sidney\footnote{\url{https://igormq.github.io/datasets/}} (SID): contains 5,777 utterances recorded by 72 speakers (20 women) from 17 to 59 years old with fields such as place of birth, age, gender, education, and occupation;
    \item Multilingual TEDx: a collection of audio recordings from TEDx talks in 8 source languages. The Portuguese set (mostly Brazilian Portuguese variant) contains 164 hours of transcribed speech. 
\end{itemize}

The assembled dataset is very similar to the base presented by  \cite{quintanilha2020open}. Besides the data used by these authors, Common Voice (CV), MLS, and Multilingual TEDx (Portuguese audios only) were also included. CSLU Spoltech \cite{neto2008spoltech} was not considered, since we opted to use only publicly  available datasets.

The majority of audios have short duration (between 1 to 10 secs), while MLS tends to have longer audios. Regarding quantity of speakers, CV is the dataset that contains the greatest amount: 2,038 in total. In contrast, MLS has only 62 speakers. Training in data with more variety of speakers is expected to lead to better results if compared to less variety. 

Some issues are present in the gathered datasets. For example, TEDx has some audios in European Portuguese and some transcription errors, MLS contains old spellings, before the last BP spelling reform, and SID has audios lacking transcriptions, acronyms instead of spelled words in transcriptions as well as arabic numbers instead of number in full. Additionally, it is expected to have some imbalances in the gathered datasets, especially VoxForge, which has the majority of its non-anonymous speakers identified as males. 

We used the original splits of the gathered data on our assembled dataset. For datasets that did not have a subdivision of their data, we created a train and a test set. Table \ref{tab:split} presents the split used for training, validating and testing the final model. Common Voice was selected as validation during training. We also augmented the Common Voice training subset, by selecting all validated instances excepting the speakers and sentences present in the original dev and test sets. Although this might insert some duplicated sentences and speakers to the final subset, it has the potential to increase the size of the dataset, as we are interested to train the model with more data as possible.

\begin{table}[]
\centering \scriptsize
\setlength{\extrarowheight}{2pt}
\caption{Dataset Split in hours. Common Voice dev was selected to validate the model.}

\begin{tabular}{l|r|r|r}    
\hline
\textbf{Dataset}               & \textbf{Train} & \textbf{Valid} & \textbf{Test} \\
\hline
CETUC                          & 93.9h              & --                  & 5.4h             \\
Common Voice*                   & 37.6h             & 8.9h                  & 9.5h            \\
LaPS BM                         & 0.8h              & --                  & 0.1h            \\
MLS                            & 161.0h             & --                  & 3.7h           \\
Multilingual TEDx (Portuguese) & 144.2h            & --                  & 1.8h           \\
SID                       & 5.0h           & --                  & 1.0h        \\
VoxForge                       & 2.8h             & --                  & 0.1h            \\
 \hline
\textbf{Total}                 & 437.2h             & 8.9h                  & 21.6h \\
\hline
\multicolumn{4}{l}{\scriptsize{*Augmented training set}}    
\end{tabular} 
\label{tab:split}
\end{table}

The test subsets were created as follows. For VoxForge, SID and LaPS BM, we selected 5\% of unique male speakers and 5\% of unique female speakers. The CETUC test set was created as proposed by \cite{quintanilha2020open}. For the remaining datasets (TEDx, MLS and Common Voice), we used the official test sets. We also performed a filtering to remove sentences of the training data that were also present in the test subsets of the final assembled dataset. This was done in order to provide unbiased training data while preserving the test one. We performed this operation on the final assembled dataset, considering all test sets, to avoid any subset contamination. For the subsets training, we do not perform any filtering besides removal of missing and empty transcriptions.

Regarding audio preprocessing, we performed some processing using Librosa\footnote{\url{https://librosa.org/}}. All audios were resampled to 16kHz. We also ignored audio with more than 30 seconds of our dataset (less than 1\% of the total). 

For the subsets experiments, we used both a NVIDIA TITAN V 12GB and a NVIDIA TESLA V100 32GB, depending on the size of the dataset used for training. The final model was trained in three NVIDIA TESLA V100 32GB. Fine-tuning parameters were defined using the same configurations of the 100-hour experiment proposed by the original Wav2vec 2.0  authors\footnote{\url{https://github.com/pytorch/fairseq/tree/master/examples/wav2vec}.}, except the number of updates and the max quantity of tokens, which was set to $10^5$ and $10^6$, respectively. As in the original work, we also used the Adam optimizer. Other parameters include: a 10k initial freeze of the transformer during fine-tuning, an augmentation similar to SpecAugment applied to time-steps and channels, a learning rate of $3 \times 10^{-5}$ and a gradient accumulation of 12 steps.
%an update frequency of 12 steps. 
The model was trained using the Fairseq\footnote{\url{https://github.com/pytorch/fairseq}} framework for a total of 117 epochs in approximately 4 days. The batch size during training was defined automatically by the framework depending on the max quantity of predefined tokens (the effective batch size was led to approximately 2,250secs of audio).

During training, the best model was selected based on the lowest WER obtained on the validation set. We do not perform validation for the subsets experiments, as we did not have validation subsets for all the gathered datasets.

We also performed experiments using language models to post-process the ASR model results following the original work of \cite{baevski2020Wav2vec}, that is, using KenLM and Transformer based LMs. In this sense, we opted to use the KenLM based language models provided by \cite{quintanilha2020open}, and also to train a new Transformer based LM \cite{baevski2018adaptive} using portuguese data.

We trained the Transformer LM using a Wikipedia based corpus\footnote{Available at \url{https://igormq.github.io/datasets}} constructed by \cite{quintanilha2020open}. This corpus is composed of over 8 million sentences extracted from the Wikipedia articles and was pre-processed removing all punctuation and tags, besides converting the numbers into their written form. We opted to use the CV validation set to validate the model while training. We trained this model for 50,000 updates using a gradient accumulation of 32 steps. We also set the max quantity of tokens equal to 1,024. Regarding hyperparameters, the predefined settings of the original repository of Fairseq\footnote{\url{https://github.com/pytorch/fairseq/blob/main/examples/language_model}} was used.

The training of the Transformer LM lasted for approximately 12 days using a NVIDIA TESLA V100 32GB GPU. The measured perplexity of the training and validation data was 29.88 and 144.6, respectively, while the measured average perplexity over the BP test sets was 195.65.

Finally, we also trained a baseline model based on DeepSpeech2 proposed by \cite{seandeepspeech} and implemented by \cite{quintanilha2020open}. This model was trained using the BP dataset and the configurations proposed by \cite{quintanilha2020open}. The model trained for 50 epochs, which seems to be adequate considering the amount of data and the results obtained by  the authors.

\section{Results and Discussion}\label{sec:results}

The final assembled dataset has approximately 470 hours of speech in total. This amount can be considered appropriate, since Wav2vec 2.0 reaches optimal results in English for experiments based on 100 and 960 hours of audio. 
As presented in Table \ref{tab:split}, 
the majority of the dataset is composed by the TEDx, MLS, followed by CETUC and Common Voice. The VoxForge and LapsBM datasets together represented less than 2\% of the total, the same for SID. In particular, LapsBM corresponds to less than 1\% of the final dataset. This means that this dataset might make little contribution to the final model. Another important aspect is the duration of the audios: the MLS dataset corresponds to approximately 40\% of the final dataset, while having only 16\% of the total number of audios present in the training set. 

% During the validation, the best model was saved considering the WER obtained in the validation set. The best model was then tested against the test sets. 
The results of this work are presented in Table~\ref{tab:results}. The lowest WER obtained in the Common Voice test set was 9.2\% using a Transformer based language model. Similar results were obtained using KenLM based models trained by \cite{quintanilha2020open}. Without LM, the WER was 14\%. Both results can be considered promising in the context of the development of ASR models given the SOTA for Brazilian Portuguese. We also tested the final model against all other test sets, obtaining 12.4\% of WER on average without LM. 

\begin{table}[]
\centering \scriptsize 
% Row extra space
\setlength{\extrarowheight}{2pt}
% Column space (default is 6pt)
% \setlength{\tabcolsep}{12pt}
\caption{Results and comparison with related works (WER) and baseline. The n-gram models are based on KenLM \cite{quintanilha2020open}.}
\begin{tabular}{l|l|l||r|r|r|r|r|r|r|r}
\hline
\multirow{2}{*}{\textbf{Experiment}} & \multirow{2}{*}{\textbf{Train data}} & \multirow{2}{*}{\textbf{LM}} & \multicolumn{7}{c}{\textbf{Test subset}} \\ \cline{4-11} 
 & & & \multicolumn{1}{l|}{\tiny \textbf{CETUC}} & \multicolumn{1}{l|}{\tiny \textbf{CV}} & \multicolumn{1}{l|}{\tiny \textbf{LaPS}} & \multicolumn{1}{l|}{\tiny \textbf{MLS}} & \multicolumn{1}{l|}{\tiny \textbf{SID}} & \multicolumn{1}{l|}{\tiny \textbf{TEDx}} & \multicolumn{1}{l|}{\tiny \textbf{VF}} & \multicolumn{1}{l}{\tiny \textbf{AVG}} \\ \hline 
Baseline & BP Dataset & No  & 0.307 & 0.444 & 0.361 & 0.442 & 0.363 & 0.552 &  0.467 & 0.419 \\  \hline 
1. All data & BP Dataset & No  & 0.052 & 0.140 & 0.074 & 0.117 & 0.121 & 0.245 & \textbf{0.118} & 0.124 \\ 
2. All data + 3-gram & BP Dataset & Yes &  0.033 & 0.095 & 0.046 & \textbf{0.123} & 0.112 & 0.212 & 0.123  & 0.106 \\ 
3. All data + 5-gram  & BP Dataset & Yes & 0.033 & 0.094 & 0.043 & \textbf{0.123} & \textbf{0.111} & \textbf{0.210} & 0.123  & \textbf{0.105} \\
4. All data + Transf.  & BP Dataset & Yes & \textbf{0.032} & \textbf{0.092} & \textbf{0.036} & 0.130 & 0.115 & 0.215 & 0.125  & 0.106 \\ \hline
5. Subsets & Subset* & No & 0.447 & 0.126 & 0.145 & 0.163 & 0.124 & 0.203 & 0.561  & 0.258 \\ 
6. Subsets + 3-gram & Subset*& Yes  & 0.333 & 0.097 & 0.073 & 0.144 & \textbf{0.104} & 0.203 & 0.453  & 0.201 \\ 
7. Subsets + 5-gram & Subset* & Yes & \textbf{0.328} & 0.096 & 0.073 & \textbf{0.143} & \textbf{0.104} & \textbf{0.201} & 0.450 & \textbf{0.199} \\
8. Subsets + Transf.  & Subset* & Yes & 0.400 & \textbf{0.086} & \textbf{0.053} & 0.165 & 0.198 & 0.204 & \textbf{ 0.406} & 0.216 \\ \hline
Batista et al. \cite{batista18iberspeech}  & Various** & Yes & - & - & 0.047 & - & - & - & -  & - \\  
Gris et al. \cite{gris2021desenvolvimento}  & LaPS BM & No & - & 0.34 & - & - & - & - & -  & - \\ 
Quintanilha et al. \cite{quintanilha2020open} & BRSD v2 & Yes  & 0.254 & - & - & - & - & - & -  & - \\ \hline
\multicolumn{9}{l}{\scriptsize{*Train subset respective to the test subset}}   \\
\multicolumn{9}{l}{\scriptsize{**No name provided}}
\end{tabular}
\label{tab:results}
\end{table}

Our more discrepant result is observed against the CETUC test set. Our trained model using only the CETUC subset achieved a WER of 32\%, while the final model obtained less than 3\% of error, when tested against the same test set. %One explanation for this result is that this dataset has a poor variety of vocabulary, since the same phrase is repeated several times by one of the 100 speakers.
One explanation for this result is that this dataset has a poor variety of  vocabulary since  the  same  phrases are repeated   by  each  of  the 100 speakers. This provides a difficult train set to generalize while the test set appears to be easier if compared to the others. A similar effect can be observed in the VF subset. In this case, a possible explanation is the dataset imbalance caused by the majority of male speakers present in the training set.

The obtained results of our final model are better than the subsets experiments, as expected. An exception is Experiment 5, which performed better than Experiment 1 against Common Voice. In this case, a small training set leads to a small error. This result is possibly  explained due to slight overfitting towards recording conditions. Similar occured to TEDx, another big dataset, but the phenomenon was not verified in smaller datasets. 

The LMs used also considerably improved the performance of the trained models. The improvement is more noticeable in the subsets experiments. In some cases, the language models appear to slightly worsen the model performance, which may be explained by the characteristics of the data used to train the language models or some particularities of each subset. The Transformer LM demonstrates better performance on CETUC, CV and LapsBM. We hypothesize that these subsets represents more the data used to train the LM, if compared to the other datasets.

% \section{Comparison with Brazilian Portuguese models}\label{sec:comparison-br}

% In the context of BP models, there are few works that we can compare results to.
In the context of BP models, there are few works that we can compare with our work. Specifically, we built a test set following \cite{quintanilha2020open} to compare this recent work with ours. Using the CV test set, we can compare the obtained results with our previous work \cite{gris2021desenvolvimento}, and using the LapsBM test subset, we can compare the results against the baseline proposed by \cite{batista18iberspeech}. The complete comparison is present in Table \ref{sec:results}. Although the baseline proposed by \cite{batista18iberspeech} is not based on deep learning, the obtained results against LapsBM seems to be notably competitive to ours. However, we believe that LapsBM does not reflect all the domains of BP data, since the results over the other test subsets diverge considerably. 
%Another important aspect to note is that experiment 5 demonstrates that less than an hour of training data (part of LapsBM) is sufficient to obtain good performance over this particular subset.

Regarding E2E models, the results from \cite{quintanilha2020open} are interesting, considering that the model training was made entirely in a supervised form. However, the result of 25.45\% of WER is above the WER of this work (3.2\%). We believe that our results were superior from the work of \cite{quintanilha2020open} for two main reasons. First, we used Wav2vec 2.0, a more modern neural architecture. Second, we trained over more data, since recently several public access dataset were released. This result is also observed with our baseline, where the model trained with BP dataset performed worse against all test sets in all experiments.

Finally, in our previous work \cite{gris2021desenvolvimento}, we obtained 34\% of WER against the CV dataset. Although we used the same model architecture as in this work, the obtained WER is worse than our current result, even without LM (14\%). The same behavior is observed in Experiment 5, in which the WER is only 12.6\%. Both results demonstrate the importance of the use of more data among in-domain training   to achieve better results.

% In a general form, our results suggest  that the unsupervised or self-supervised techniques can be better alternatives for the development of ASR models for Brazilian Portuguese, especially  while the pure supervised techniques need a great quantity of labeled data to generalize as well as the English ones. The obtained results also suggest  that our model achieves SOTA performance for the Brazilian Portuguese, considering public available models.

\section{Conclusions}\label{sec:conc}

In this work we presented a model for Automatic Speech Recognition for the Brazilian Portuguese Language. Our results show that self-supervised learning is a great advance in the development of ASR systems for BP, mainly because it requires less labeled data for training, achieving a better performance if compared to the models trained entirely on a supervised form. On average, our model obtained 10.5\% and 12.4\% of WER against the proposed test sets, with and without a language model, respectively. According to our knowledge, the model achieves state-of-the-art results among the open available E2E models for the target language. 

In addition, our work suggests that the use of datasets with a large variety of vocabulary and speakers are still important to the development of robust models, as well as in-domain training. Besides, it is not clear how much these models are sensitive to noisy data. We also do not investigate the development of robust LM models, which can further improve the results. In this sense, as future work, we plan to train more robusts models with new datasets and to investigate the use of data augmentation techniques, such as additive noise \cite{snyder2018x} and Room Impulse Response (RIR) simulation \cite{ko2017study}. We believe that these approaches can collaborate with the community towards the development of robust ASR models for Brazilian Portuguese.

\section*{Acknowledgements}
\small{
This research was funded by CEIA with support by the Goi\'{a}s State Foundation (FAPEG grant \#201910267000527)\footnote{\url{http://centrodeia.org/}}, Department of Higher Education of the Ministry of Education (SESU/MEC), Copel Holding S.A.\footnote{\url{https://www.copel.com}}, and Cyberlabs Group\footnote{\url{https://cyberlabs.ai/}}.  Also, this study was financed in part by the Coordena\c{c}\~{a}o de Aperfei\c{c}oamento de Pessoal de N\'{i}vel Superior -- Brasil (CAPES) -- Finance Code 001. We also would like to thank Nvidia Corporation for the donation of Titan V GPU used in part of the experiments presented in this research.
}

\newpage
\bibliographystyle{splncs04}
\bibliography{referencias}

\begin{thebibliography}{10}
\providecommand{\url}[1]{\texttt{#1}}
\providecommand{\urlprefix}{URL }
\providecommand{\doi}[1]{https://doi.org/#1}

\bibitem{AGUIARDELIMA2020101055}
{Aguiar de Lima}, T., {Da Costa-Abreu}, M.: A survey on automatic speech
  recognition systems for portuguese language and its variations. Computer
  Speech \& Language  \textbf{62},  101055 (2020).
  \doi{https://doi.org/10.1016/j.csl.2019.101055},
  \url{https://www.sciencedirect.com/science/article/pii/S0885230819302992}

\bibitem{alencar2008lsf}
Alencar, V., Alcaim, A.: Lsf and lpc-derived features for large vocabulary
  distributed continuous speech recognition in brazilian portuguese. In: 2008
  42nd Asilomar Conference on Signals, Systems and Computers. pp. 1237--1241.
  IEEE (2008)

\bibitem{amodei2016deep}
Amodei, D., Ananthanarayanan, S., Anubhai, R., Bai, J., Battenberg, E., Case,
  C., Casper, J., Catanzaro, B., Cheng, Q., Chen, G., et~al.: Deep speech 2:
  End-to-end speech recognition in english and mandarin. In: International
  conference on machine learning. pp. 173--182. PMLR (2016)

\bibitem{ba2016layer}
Ba, J.L., Kiros, J.R., Hinton, G.E.: Layer normalization. arXiv preprint
  arXiv:1607.06450  (2016)

\bibitem{baevski2018adaptive}
Baevski, A., Auli, M.: Adaptive input representations for neural language
  modeling. In: International Conference on Learning Representations (2018)

\bibitem{baevski2019vq}
Baevski, A., Schneider, S., Auli, M.: vq-wav2vec: Self-supervised learning of
  discrete speech representations. In: International Conference on Learning
  Representations (ICLR) (2020), \url{https://openreview.net/pdf?id=rylwJxrYDS}

\bibitem{baevski2020Wav2vec}
Baevski, A., Zhou, Y., Mohamed, A., Auli, M.: wav2vec 2.0: A framework for
  self-supervised learning of speech representations. In: Larochelle, H.,
  Ranzato, M., Hadsell, R., Balcan, M.F., Lin, H. (eds.) Advances in Neural
  Information Processing Systems. vol.~33, pp. 12449--12460. Curran Associates,
  Inc. (2020),
  \url{https://proceedings.neurips.cc/paper/2020/file/92d1e1eb1cd6f9fba3227870bb6d7f07-Paper.pdf}

\bibitem{batista18iberspeech}
Batista, C., Dias, A.L., {Sampaio Neto}, N.: {Baseline Acoustic Models for
  Brazilian Portuguese Using Kaldi Tools}. In: Proc. IberSPEECH 2018. pp.
  77--81 (2018). \doi{10.21437/IberSPEECH.2018-17}

\bibitem{conneau2020unsupervised}
Conneau, A., Khandelwal, K., Goyal, N., Chaudhary, V., Wenzek, G., Guzm{\'a}n,
  F., Grave, {\'E}., Ott, M., Zettlemoyer, L., Stoyanov, V.: Unsupervised
  cross-lingual representation learning at scale. In: Proceedings of the 58th
  Annual Meeting of the Association for Computational Linguistics. pp.
  8440--8451 (2020)

\bibitem{davis1952automatic}
Davis, K.H., Biddulph, R., Balashek, S.: Automatic recognition of spoken
  digits. The Journal of the Acoustical Society of America  \textbf{24}(6),
  637--642 (1952)

\bibitem{goodfellow}
Goodfellow, I., Bengio, Y., Courville, A.: Deep learning. MIT press (2016)

\bibitem{graves2006connectionist}
Graves, A., Fern{\'a}ndez, S., Gomez, F., Schmidhuber, J.: Connectionist
  temporal classification: labelling unsegmented sequence data with recurrent
  neural networks. In: Proceedings of the 23rd international conference on
  Machine learning. pp. 369--376 (2006)

\bibitem{gris2021desenvolvimento}
Gris, L.R.S., Casanova, E., de~Oliveira, F.S., da~Silva~Soares, A.,
  Candido-Junior, A.: Desenvolvimento de um modelo de reconhecimento de voz
  para o portugu{\^e}s brasileiro com poucos dados utilizando o wav2vec 2.0.
  In: Anais do XV Brazilian e-Science Workshop. pp. 129--136. SBC (2021)

\bibitem{heafield2011kenlm}
Heafield, K.: Kenlm: Faster and smaller language model queries. In: Proceedings
  of the sixth workshop on statistical machine translation. pp. 187--197 (2011)

\bibitem{hendrycks2020gaussian}
Hendrycks, D., Gimpel, K.: Gaussian error linear units (gelus) (2020)

\bibitem{hochreiter1997long}
Hochreiter, S., Schmidhuber, J.: Long short-term memory. Neural computation
  \textbf{9}(8),  1735--1780 (1997)

\bibitem{karpagavalli2016review}
Karpagavalli, S., Chandra, E.: A review on automatic speech recognition
  architecture and approaches. International Journal of Signal Processing,
  Image Processing and Pattern Recognition  \textbf{9}(4),  393--404 (2016)

\bibitem{ko2017study}
Ko, T., Peddinti, V., Povey, D., Seltzer, M.L., Khudanpur, S.: A study on data
  augmentation of reverberant speech for robust speech recognition. In: 2017
  IEEE International Conference on Acoustics, Speech and Signal Processing
  (ICASSP). pp. 5220--5224. IEEE (2017)

\bibitem{li2021recent}
Li, J.: Recent advances in end-to-end automatic speech recognition. arXiv
  preprint arXiv:2111.01690  (2021)

\bibitem{seandeepspeech}
Naren, S.: Speech recognition using deepspeech2 (2020),
  \url{https://github.com/SeanNaren/deepspeech.pytorch}

\bibitem{neto2011free}
Neto, N., Patrick, C., Klautau, A., Trancoso, I.: Free tools and resources for
  brazilian portuguese speech recognition. Journal of the Brazilian Computer
  Society  \textbf{17}(1),  53--68 (2011)

\bibitem{neto2008spoltech}
Neto, N., Silva, P., Klautau, A., Adami, A.: Spoltech and ogi-22 baseline
  systems for speech recognition in brazilian portuguese. In: International
  Conference on Computational Processing of the Portuguese Language. pp.
  256--259. Springer (2008)

\bibitem{Park_2019}
Park, D.S., Chan, W., Zhang, Y., Chiu, C.C., Zoph, B., Cubuk, E.D., Le, Q.V.:
  Specaugment: A simple data augmentation method for automatic speech
  recognition. Interspeech 2019  (Sep 2019).
  \doi{10.21437/interspeech.2019-2680},
  \url{http://dx.doi.org/10.21437/Interspeech.2019-2680}

\bibitem{povey2011kaldi}
Povey, D., Ghoshal, A., Boulianne, G., Burget, L., Glembek, O., Goel, N.,
  Hannemann, M., Motlicek, P., Qian, Y., Schwarz, P., et~al.: The kaldi speech
  recognition toolkit. In: IEEE 2011 workshop on automatic speech recognition
  and understanding. No.~CONF, IEEE Signal Processing Society (2011)

\bibitem{Pratap_2020}
Pratap, V., Xu, Q., Sriram, A., Synnaeve, G., Collobert, R.: Mls: A large-scale
  multilingual dataset for speech research. Interspeech 2020  (Oct 2020).
  \doi{10.21437/interspeech.2020-2826},
  \url{http://dx.doi.org/10.21437/Interspeech.2020-2826}

\bibitem{quintanilha2017end}
Quintanilha, I.M.: End-to-end speech recognition applied to brazilian
  portuguese using deep learning. MSc dissertation  (2017)

\bibitem{quintanilha2020open}
Quintanilha, I.M., Netto, S.L., Biscainho, L.W.P.: An open-source end-to-end
  asr system for brazilian portuguese using dnns built from newly assembled
  corpora. Journal of Communication and Information Systems  \textbf{35}(1),
  230--242 (2020)

\bibitem{schneider2019wav2vec}
Schneider, S., Baevski, A., Collobert, R., Auli, M.: wav2vec: Unsupervised
  pre-training for speech recognition. In: INTERSPEECH (2019)

\bibitem{schramm2006cslu}
Schramm, M., Freitas, L., Zanuz, A., Barone, D.: Cslu: Spoltech brazilian
  portuguese version 1.0 ldc2006s16 (2006)

\bibitem{snyder2018x}
Snyder, D., Garcia-Romero, D., Sell, G., Povey, D., Khudanpur, S.: X-vectors:
  Robust dnn embeddings for speaker recognition. In: 2018 IEEE International
  Conference on Acoustics, Speech and Signal Processing (ICASSP). pp.
  5329--5333. IEEE (2018)

\bibitem{vaswani2017attention}
Vaswani, A., Shazeer, N., Parmar, N., Uszkoreit, J., Jones, L., Gomez, A.N.,
  Kaiser, L., Polosukhin, I.: Attention is all you need. In: Neural Information
  Processing Systems (NIPS) (2017)

\bibitem{yu2016automatic}
Yu, D., Deng, L.: Automatic Speech Recognition. Springer (2016)

\end{thebibliography}

\end{document}